\documentclass[10pt,journal,compsoc]{IEEEtran}
\usepackage{epstopdf}

\usepackage{graphicx}
\usepackage{subfigure}
\usepackage{amsmath,amssymb}
\usepackage{multirow}
\usepackage{float}
\usepackage{CJK}
\usepackage{indentfirst}
\usepackage{bm}
\usepackage{setspace}
\usepackage{lineno,hyperref}
\usepackage{booktabs}
\usepackage{caption}
\ifCLASSOPTIONcompsoc
  \usepackage[nocompress]{cite}
\else
  \usepackage{cite}
\fi

\ifCLASSINFOpdf
\else
\fi

\begin{document}
%
\title{Knowledge-guided Semantic Computing Network}

\author{ Guangming~Shi,~\IEEEmembership{Senior~Member,~IEEE}
    Zhongqiang~Zhang,
    Dahua~Gao$^*$,
	Xuemei~Xie,~\IEEEmembership{Member,~IEEE,}
	Yihao~Feng,
    Xinrui~Ma,
	and Danhua~Liu
	\IEEEcompsocitemizethanks{\IEEEcompsocthanksitem Guangming Shi, Zhongqiang Zhang, Dahua Gao, Xuemei Xie, Yihao Feng, Xinrui Ma, Danhua Liu are with School of Artificial Intelligence, Xidian University, Xi'an, Shanxi, 710071, China.\protect\\
\IEEEcompsocthanksitem $^*$ Corresponding author: Dahua Gao. E-mail: gaodahua@gmail.com }}
		


%

\IEEEtitleabstractindextext{%
\begin{abstract}
It is very useful to integrate human knowledge and experience into traditional neural networks for faster learning speed, fewer training samples and better interpretability.  However, due to the obscured and indescribable black box model of neural networks, it is very difficult to design its architecture, interpret its features and predict its performance. Inspired by human visual cognition process, we propose a knowledge-guided semantic computing network which includes two modules: a knowledge-guided semantic tree and a data-driven neural network. The semantic tree is pre-defined to describe the spatial structural relations of different semantics, which just corresponds to the tree-like description of objects based on human knowledge. The object recognition process through the semantic tree only needs simple forward computing without training. Besides, to enhance the recognition ability of the semantic tree in aspects of the diversity, randomicity and variability, we use the traditional neural network to aid the semantic tree to learn some indescribable features. Only in this case, the training process is needed. The experimental results on MNIST and GTSRB datasets show that compared with the traditional data-driven network, our proposed semantic computing network can achieve better performance with fewer training samples and lower computational complexity. Especially, Our model also has better adversarial robustness than traditional neural network with the help of human knowledge.
\end{abstract}

}

\maketitle

\IEEEdisplaynontitleabstractindextext

\IEEEpeerreviewmaketitle

\IEEEraisesectionheading{\section{Introduction}\label{sec:introduction}}

\IEEEPARstart{E}{Recently}, a series of deep learning algorithms including CNN\cite{lecun1998gradient}, GAN\cite{goodfellow2014generative}, CapsNet\cite{sabour2017dynamic} etc. have made remarkable contributions in the field of computer vision such as image recognition\cite{krizhevsky2012imagenet}, object detection\cite{ren2015faster} and image caption\cite{vinyals2015show}. However, such neural networks are trained in a big-data-driven approach, which results in a series of disadvantages including high energy consumption, large storage space requirements, cumbersome manual annotation processes and data acquisition difficulties. What is more, they are also suffered seriously from adversarial attacks.

Furthermore, unlike humans, most traditional neural networks have an unsatisfactory performance on few-shot learning because of too many features to learn and difficulties in reasoning through prior knowledge. A lot of researches show that human can learn new concepts and generalize meaningfully from just few positive examples. Even children can make meaningful generalizations via 'one-shot learning' \cite{xu2007word}. An influential opinion explains that during the new task learning, human can use strong previous prior knowledge\cite{clark1987principle} to make relevant inferences. This enables us to eliminate some unreasonable results directly and accelerate our learning process. Thus, integrating human knowledge into neural networks is an effective solution to few-shot and fast learning. However, due to the indescribable black box model of the traditional neural network, it is very difficult to design its architecture, interpret its features and predict its performance, resulting in difficulties of the integration. Therefore, the existing models still need further improvement.

Fortunately, the visual cognition process of human brain ~\cite{Riesenhuber1999Hierarchical} provides a very good idea for this improvement. After visual information is transmitted to the brain via the eyes, it is processed by the human visual cortex. The visual cortex mainly includes the primary visual cortex (V1, also known as the striate cortex) and the extrastriate cortex (such as V2, V3, V4, V5, etc.). In 1962, Hubel and Wiesel\cite{hubel1962receptive} found that some cells in V1 only respond to bright or dark strips with special orientations and for each cell, there is an optimal position in which the cell reacts most strongly. Riesenhuber M et al ~\cite{Riesenhuber1999Hierarchical} describe a new hierarchical model consistent with physiological data from inferior temporal cortex that accounts for this complex visual task and makes testable predictions. Matthew Lawlor et al points out that long-range horizontal connections among V1 cells enable V1 to respond to curvature \cite{lawlor2014feedforward}. Besides, Livingstone and Hubel\cite{livingstone1988segregation} proposed that different types of V1 cells make up three different structures in V1, which respectively perceives and transmits the information about the color, the shape, the movement and stereoscopic vision to V2 and subsequent cells. During the whole human visual cognition process, visual information goes through two different visual pathways. One of them called 'dorsal stream' leads to the parietal cortex for spatial vision (the information about 'where'); another one called 'ventral stream' leads to inferior temporal cortex for object vision (the information about 'what')\cite{ungerleider1994and}.

Areas along both pathways are hierarchical structures, such that low-level inputs are transformed into more useful representations through successive stages of processing. In this process, average receptive field size increases steadily and neuronal response properties become increasingly complex. For example, along the 'ventral stream' for object vision, V1, V2 and some V4 cells function as local filters and respond to low-level features including tiny strips, contours and colors; and inferior temporal cells respond selectively to global or overall object features, such as shape\cite{ungerleider1994and}; after that, discrete cortical regions respond preferentially to specific objects such as faces, buildings, letters, houses, etc.\cite{ishai1999distributed}. And then the information is further analyzed and integrated to become the final perceived vision.

This hierarchical transmission and formation process of visual information in human brain gives researchers great inspiration. In recent years, there have been many studies on building recognition networks from the perspective of the hierarchical composition of objects. S. C. Zhu et al. propose And-Or Templates (AOT) for object recognition and detection\cite{zhu2007stochastic,si2013learning}. It is a hierarchical reconfigurable image template and learned the structural semantic composition of objects through probabilistic methods. Dileep Georage et al proposes a recursive cortical network which is a generative model combining AOT with conditional random fields \cite{george2017generative}. It breaks the defense of text-based captchas and is more efficient than neural networks on scene text recognition. Furthermore, Hinton proposes CapsNet \cite{sabour2017dynamic, e2018matrix} to deal with the representation of spatial hierarchicy between simple object and complex object, which is a problem for CNN. The CapsNet uses a vector or matrix called capsule to represent a group of neurons with different properties. Each capsule can be considered as a specific type of entity such as an object or an object part, whose length represents the probability that a capsule's entity exists. This makes CapsNet achieve excellent performance on MNIST and recognize highly overlapping digits.

Previous researches make some preliminary studies on imitating the human decision-making and visual cognition process. But there is no such thing as a free lunch. The corresponding networks become more complex with the requirement of massive data and the long training process because they are still essentially data-driven methods. So far, there are still no efficient mechanisms to utilize the previous experience and knowledge of human. Therefore, it is essential to exploit a way to integrate human knowledge and experience into traditional neural networks for faster learning speed, fewer training samples and better interpretability. Fortunately, human beings have intuitive and hierarchical structural cognitive abilities according to previous researches about human visual cognition process. Therefore the thinking way of the human brain is a kind of intuitive thinking. In this paper, we model such a thinking way with the semantic tree, which is a tree-like structure with semantic dictionary and semantic relation template. We propose a novel recognition approach for classification tasks called the semantic computing network (SCN). The SCN consists of two modules: a knowledge-guided semantic tree module and a data-driven traditional neural network module. In order to build the semantic tree module, we firstly use the human knowledge to construct semantics hierarchically with a semantic dictionary. Then we define several spatial structural relations between semantics based on experience and get a set of semantic relation templates through statistical method. Finally, we build the semantic tree for object recognition based on the semantic dictionary and the relation templates. In the recognition process, semantic primitive detection is performed on the input images, and then the semantic tree gives a recognition result through the forward computing. The kind of semantic tree is based on knowledge and can be interpreted, expanded easily. But it is hard for human to fully represent the specially complex images with the amount of details, thus we need to represent complex details with the help of traditional neural network. In our method, we choose the CapsNet as an auxiliary leaning-based network because of its high compatibility with the semantic tree. Through the combination of the knowledge-guided semantic tree and the data-driven CapsNet, the SCN keeps a high recognition accuracy with fewer training samples, lower computational complexity and better interpretability. At the same time, our SCN has more robustness to adversarial attacks than traditional neural networks.

Overall, the contributions of this study are mainly in three aspects:

\begin{itemize}
\item We firstly exploit a way to model human visual cognition process with semantic tree. This hierarchical semantic tree have better description and interpretation to the object. At the same time, it can quickly calculate semantic results.
\item We propose a knowledge-guided semantic compute network (SCN), which includes two modules: a knowledge-guided semantic tree and a data-driven CapsNet. Such SCN can achieve better performance than traditional neural network on different datasets and corresponding fewer training samples. At the same time, our SCN only needs lower computational complexity. What is more, Our SCN has better robustness to adversarial attacks than traditional neural networks.
\item We design a novel loss function for SCN, which can automatically adjust the weights between semantic tree and neural network according to the recognition ability
    of semantic tree on training samples. Such weights reflects the guidance of semantic tree (human knowledge) to traditional neural network. Finally, the traditional neural network can aid semantic tree to learn some indescribable featrues.

\end{itemize}

\section{Related Works}

Recently, with the development of traditional neural networks, more and more people pay attention to the problems of network interpretability, fewer samples, network security and network simplification.

In general, most of neural networks achieve a high accuracy at the cost of low interpretability of black-box representations, which makes it difficult for researchers to design and regulate network architecture ~\cite{zhang2018visual}. In order to understand the decision-making process of traditional neural network from the perspective of visual cognition process, semantic interpretability is a critical research direction. Some explainable models recently have been built: some researchers explore the way to combine features visualization and other interpretable techniques to understand how traditional neural networks make decisions\cite{olah2018building}; Zhang et al. make a review of CNN interpretability research, which gives more details about interpretability researches of CNN \cite{zhang2017interpretable}; Wu et al. present a method of learning qualitatively interpretable models in object detection ~\cite{DBLP:journals/corr/abs-1711-05226}. Sara Sabour, Nicholas Frosst and Geoffrey E Hinton et al design a novel CapsNet ~\cite{sabour2017dynamic}, which has a certain interpretability. We find its structure is similar with human visual recognition process.  These researches motivate us to imitate the human decision-making process by designing a novel network architecture.

In order to reduce training samples, There are a large number of literatures in few-shot learning ~\cite{xu2007word, pmlr-v48-santoro16, DBLP:journals/corr/KaiserNRB17, NIPS2017_6996}. Sung et al present a conceptually simple, flexible, and general framework for few-shot learning ~\cite{sung2017learning}. F.Li et al explore a bayesian method to learn much information about a category from just one or some images ~\cite{1597116}. Oriol Vinyals propose a network that maps a small labelled support set and an unlabelled example to its label. There is no need for fine-tuning to adapt to new class types ~\cite{NIPS2016_6385}

Some researchers are becoming more and more interested in the more compact and simple networks, Han et al introduce a "deep compression" method ~\cite{DBLP:journals/corr/HanMD15}, which needs three stage pipeline: pruning, trained quantization and Huffman coding. Landola et al propose a small DNN architecture called SqueezeNet, which achieves AlexNet-level accuracy on ImageNet with 50 times fewer parameters  ~\cite{DBLP:journals/corr/IandolaMAHDK16}.

Recently, some researchers pay more and more attention to the security problem of neural networks ~\cite{Barreno2010The, Barreno2006Can}. Marco Barreno et al present a framework for analyzing attacks on machine learning systems and defense against them. Some researches also find that most traditional neural networks are very vulnerable to adversarial attacks ~\cite{DBLP:journals/corr/KurakinGB16, DBLP:journals/corr/abs-1710-08864, Papernot:2017:PBA:3052973.3053009, DBLP:journals/corr/HuangPGDA17, Narodytska2017Simple, DBLP:journals/corr/SzegedyZSBEGF13}. C. Szegedy et al find that they can make the neural network classify incorrectly to an image by applying a certain imperceptible perturbation. A. Fawzi et al provide a theoretical framework for analyzing the robustness of classifiers to adversarial perturbations and show fundamental upper bounds on the robustness of classifiers ~\cite{Fawzi2015Analysis}. An adversarial sample is the input data which has been modified very slightly in some way. Although human beings can not notice any modification, neural network can easily misclassify such an adversarial sample. FGSM ~\cite{Goodfellow2014Explaining} method is regarded as a simple and straightforward strategy for generating adversarial samples, which can drastically decrease accuracy in convolutional neural networks on image classification tasks. In the same way, BIM ~\cite{Goodfellow2014Explaining} method is a more complex adversarial attack, which simply takes multiple smaller steps in FGSM attack to create the adversarial samples.

Generally, these characteristics are also reflected in the process of human recognition mechanism. Human visual recognition mechanism can explain object recognition process of traditional neural networks, which is also a simple forward computing process. Human knowledge has a strong generalization ability, and it can learn very well on fewer samples.
Human visual recognition mechanism has a very stong robustness. Due to these advantages, Next we will explore how to integrate human knowledge into the existing neural networks.

The rest of this paper is organized as follows: In Section 3, we briefly introduce the semantic computing network and some related concepts. In Section 4, we design the network architecture of the SCN and describe its working mechanism. In Section 5, we make a series of experimental evaluations on the MNIST, GTSRB datasets, corresponding few samples and adversarial samples. Finally we conclude the paper and look ahead to future work in Section 6.

\section{Semantic computing network}
\begin{figure}[ht]
  \centering
  \includegraphics [width=0.5\textwidth] {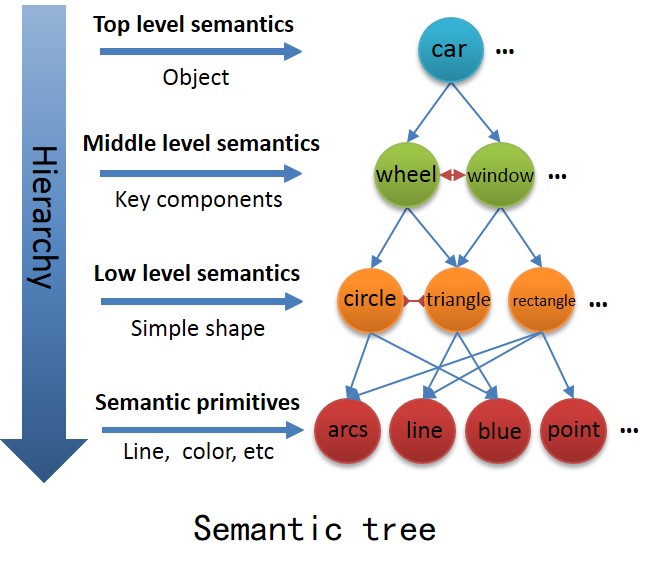}\\
  \caption{ A hierarchical semantic tree constructed by human knowledge and experience.}
  \label{fig1}
\end{figure}

In this paper, semantics generally represent some spatial structural units that have specific meaning, which consist of concrete semantics and abstract semantics. Concrete semantics refer to the objective things that can be perceived by human while abstract semantics refer to the subjective concepts generalized by human. And this paper only discusses concrete semantics. For human visual recognition process, we consider the semantic as a meaningful part in the image. And it is a kind of hierarchy, which means that a semantic can be decomposed into many sub-semantics (Fig. \ref{fig1}). For example, an image of rabbit contains the top-level semantic of 'rabbit' and it can be decomposed into rabbit's head, rabbit's leg and other sub-semantics, and the head can be further decomposed into eyes, nose, ears, etc. The semantic is decomposed into semantic primitives layer by layer in such a human visual recognition process approach. The so-called semantic primitives are the the most typical basic units of semantics as well as basic visual perceptions of human such as strips and colors which V1 cells can perceive.

In order to represent human knowledge, we construct a semantic dictionary in a hierarchical manner. Its top layers store the highest level of semantics, which represent the objects. The middle layers store sub-semantics (the parts of objects) and the bottom layers store a large number of semantic primitives. In this paper, we use a standard image template to represent each semantic, in this way, the semantic dictionary serves as a memory bank of the network for further use. Actually, the semantic dictionary forms a mapping from semantic primitives to objects through this hierarchical structure. In the semantic dictionary, each semantic is independent and different upper semantics may have the same sub-semantics. This means the semantic dictionary can be easily expanded and sub-semantics can be shared by different upper semantics to save space and improve efficiency.

Then we define a set of structural relations among same-level semantics, including distance relations, location relations, etc. Furthermore we set up semantic relation templates for each upper-semantic through template matching and statistic method. Such semantic relation templates describe how sub-semantics compose upper semantics. Because the tree-like structure can well describe the parent-child relations between the upper and lower semantics, we build a semantic tree based on the semantic dictionary and the semantic relation templates. The semantic tree can easily complete the recognition process only via a forward computing method, which includes detecting different semantic primitives in the image, calculating the semantic relation table and finding the most suitable semantic relation template. Based on the introduction of the semantic dictionary, the recognition process has changed from data-driven to knowledge-driven. This allows such recognition to be easily understood by humans, unlike CNN's black box model. In general, we use this semantic tree to model human visual recognition process.

\begin{figure*}[ht]
  \centering
  \includegraphics [width=1\textwidth] {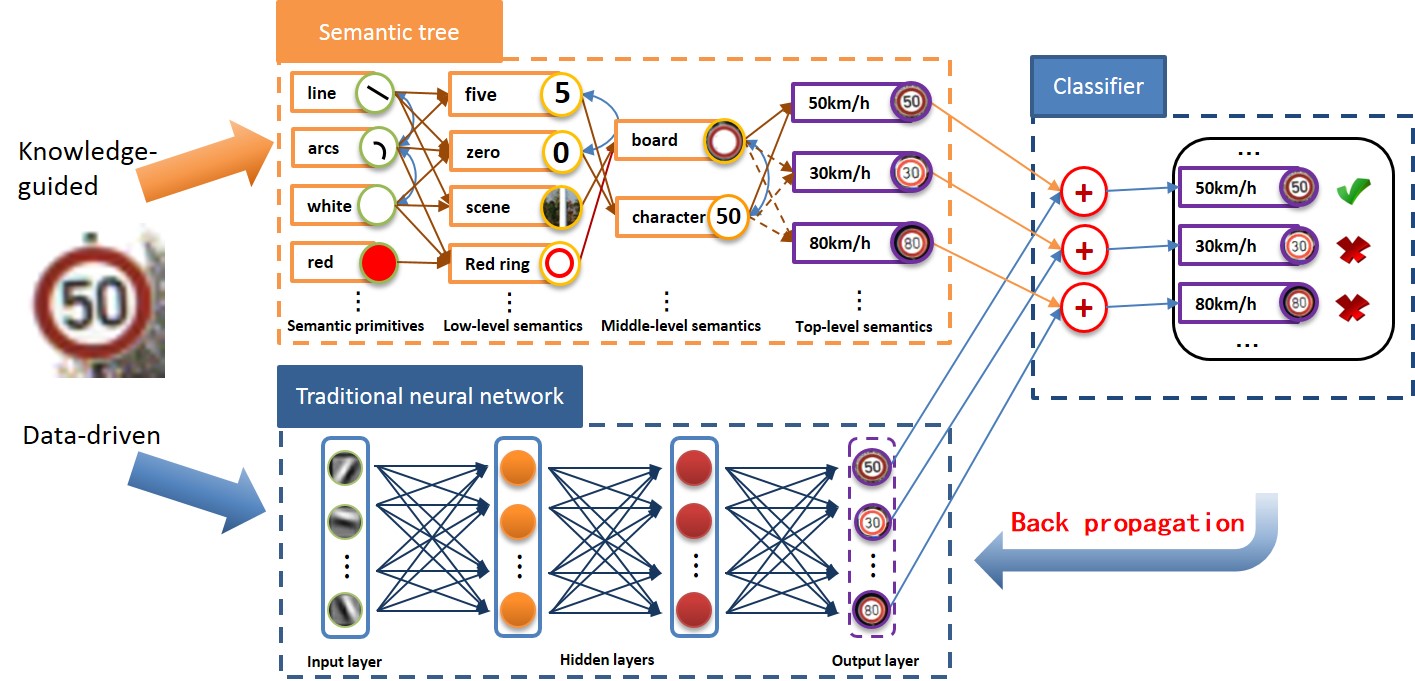}\\
  \caption{ Semantic compute network includes two modules: the knowledge-driven semantic tree module and data-driven deep neural network module. Especially, back propagation is only used to update the parameters of deep neural network.}
  \label{fig2}
\end{figure*}

As shown in Fig.\ref{fig1}, we use semantic tree to represent target knowledge and take a car as an example here. The object car is regarded as top level semantic, it can be decomposed into middle level semantics(key components) such as wheel, window, etc. Furthermore, The middle level semantics also can be decomposed into low level semantics (simple shape) such as circle, triangle, rectangle, etc. In this way, the decomposition of the semantic layer by layer until the last semantic primitives such as arcs, line, blue, etc. This semantic tree is built from primitives to system, from local to whole. Especially, layer-to-layer connections use AND or logical operations and the spatial relations are used to represent the same layer lateral connections. The benefits of this semantic tree are mainly descriptive and understandable. So it is very useful to represent object structure information with semantic tree.

However, most objects in the datasets have a variety of details such as postures, and even the background is extremely complicated. While human representation of knowledge is methodic and limited. Some complex features are hard to be described by human knowledge. Therefore this makes it inadequate to describe all features of objects through the semantic tree. However, learning-based neural networks can learn the unique representation of complex features of objects, including features that can not be described by humans. Thus, we need to search for a learning-based approach to aid the semantic tree in learning the indescribable features. In this paper, we finally choose the CapsNet as an auxiliary network because the concepts of capsules and semantics are corresponding to each other. The CapsNet is highly compatible with our semantic tree in some aspects and the two modules can be combined easily. First of all, they are both hierarchical structures. Low-level capsules correspond to 'parts' while high-level capsules correspond to 'objects', which is exactly similar with low-level semantics and high-level semantics. Secondly, the CapsNet uses an activity vector to represent the instantiation parameters of a specific type of entity \cite{sabour2017dynamic}. Similarly, the semantic tree uses vectors to describe the semantic primitives and their relations. Finally, the recognition process of two modules is same, and they both complete the classification based on the entities' score.

As shown in Fig.\ref{fig2}, We design a semantic computing network (SCN) by combining the semantic tree and the traditional neural network. Here, we take a 50km/h traffic sign of GTSRB dataset as an example. When it is input to the knowledge-guided semantic tree, we can obtain semantic primitives including line, arcs, white, red. The two way blue arrow represents spatial structure relations. Furthermore, we use simple shapes such as five, zero, scene, red ring as low-level semantics. We obtain middle level semantics such as character 50 and board according to the relation of different low-level semantics. Finally, we obtain top semantics such as 50km/h, 30km/h, 80km/h traffic signs. When object image is input to the data-driven traditional neural network, we can obtain the final output probability through the input layer, hidden layers and output layer.
The semantic tree completes preliminary and principal recognition of objects according to human knowledge and statistic characteristics. And the neural network learns the representation of indescribable features to supplement the semantic tree. Finally, we make a proportional fusion of the two module's outputs. In this way, the recognition process is completed in a knowledge-guided and learning-aided approach. Especially, semantic tree only needs simple forward computing and the back propagation ~\cite{Plaut1986Experiments} is only used to update the parameters of traditional neural network. Most important of all, the proposed network keeps a very high recognition accuracy and requires fewer training samples, smaller storage space and lower energy consumption. In the next section, we will detail the structure of the SCN in some specific applications.

\section{Network Architecture: semantic tree and the CapsNet}

\begin{figure*}[ht]
  \centering
  \includegraphics [width=1\textwidth] {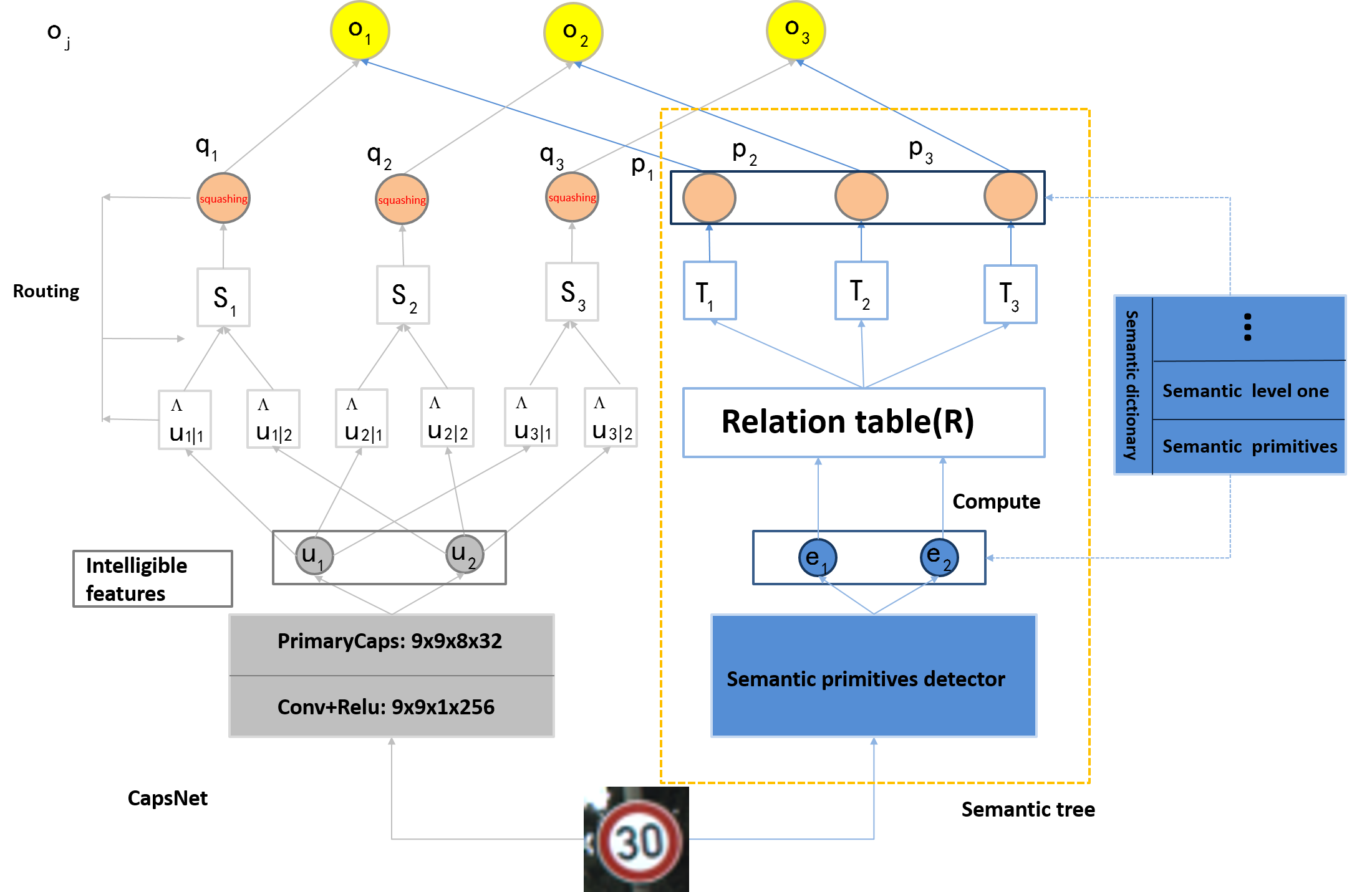}\\
  \caption{ A simple semantic computing network with two modules. One is the semantic tree module built through prior knowledge, the other is the CapsNet module acted as an assisted network.}
  \label{fig3}
\end{figure*}

In this paper, we propose a novel semantic computing network for classification tasks, which includes a semantic tree module and an aided CapsNet module. In particular, the semantic tree module completes preliminary recognition of objects according to pre-define and statistic priors. And the CapsNet module is used to learn some features that are hard to be described by the human knowledge (Fig. \ref{fig4}(b)). In this architecture, the CapsNet aids the semantic tree for better recognition performance by representing indescribable features. This architecture can be easily expanded, and we take the structure of one layer with three classes as an example to show its working mechanism (see Fig. \ref{fig3}).

As shown in Fig. \ref{fig3}, the semantic tree module includes a semantic primitive detector, a semantic dictionary, a relation table $\textbf{R}$ and several semantic relation templates $\textbf{T}$.  The CapsNet module includes the traditional convolutional layer, PrimaryCaps layer and DigitCaps layer. When inputting an image to the the SCN, we use the length of the upper capsule to represent output probability of the CapsNet module \cite{sabour2017dynamic}. In the semantic tree module, we obtain corresponding semantic primitive$(e_{1},e_{2})$ vectors through the semantic primitive detector. Then, the semantic relation table $\textbf{R}$ can be obtained through computing the relation value among semantic primitives. And the final output probability $p_{j}$ can be achieved through the product operation between relation tables $\textbf{R}$ and semantic relation templates $\textbf{T}$. Finally, we make a proportional fusion of the two module's outputs. The s$softmax$ classifier is used to obtain the output probabilities of the SCN.

\subsection{Semantic primitive detection and semantic relation table}

Given an image, we choose basic graphical shapes, such as straight line, circle, ellipse, arc, triangle, rectangular, quadrilateral, polygon and their color, etc as the semantic primitives. In this paper, we adopt the standard template matching method to detect semantic primitives. During this process, we can also obtain some corresponding properties of semantic primitives such as quantity and length.

After we find out all semantic primitives of the image, next task is to describe how the semantic primitives make up the upper semantics.For this purpose, we propose four relation tables to describe structural relations of the semantic primitives: intersect relation table, distance relation table, included-angle relation table, and location relation table. These four relation tables are abbreviated to int, dis, ang, loc respectively. Int describes whether two primitives intersect. Dis describes the distance among the primitive points¡¯ coordinates. Ang are the included-angle relations of different semantic primitives. Loc contains up, down, left and right relations.  Here, we take a location relation table as an example (Table \ref{tab:table1}). These relation tables can be achieved by direct computing. Each element in relation tables represents the relation value between two specific semantic primitives and we denote $r^{\text{c}}_{i}$  as the i-th relation value in relation table $\text{c}$, where $\text{c} \in \{\text{int}, \text{dis}, \text{ang}, \text{loc} \}$. For an image, if it has the corresponding semantic primitives and relation, $r^{\text{c}}_{i}$ =1; otherwise, $r^{\text{c}}_{i}$ =0. For example, whether the distance between two specific semantic primitives is above the threshold set in light of prior knowledge.
Then, the relation table $\textbf{R}^{\text{c}}$ can be expressed as
$
 \textbf{R}^{\text{c}} = (r^{\text{c}}_{1},r^{\text{c}}_{2}\ldots r^{\text{c}}_{\alpha_{\text{c}}}),
$
where $\alpha_{\text{c}}$  represents the dimension of relation table ${\text{c}}$.
The corresponding semantic relation table is
\begin{equation}
\label{con:f1}
 \textbf{R} = (\textbf R^{\text{int}},\textbf R^{\text{dis}},\textbf R^{\text{ang}}, \textbf R^{\text{loc}})
\end{equation}

\begin{table}[ht]
\centering
\caption{Location relation table}
\label{tab:table1}
  \begin{tabular}{ccccc}
    \toprule
     relations & primitive 1 & primitive 2 & primitive 3 & $\cdots$ \\
    \midrule
     primitive 1 & empty & down & right & $\cdots$ \\
     primitive 2 & up & empty & outside & $\cdots$ \\
     primitive 3 & left & inside & empty & $\cdots$ \\
     $\vdots$ & $\vdots$ & $\vdots$ & $\vdots$ &  $\ddots$ \\
    \bottomrule
  \end{tabular}
\end{table}

As shown in Table \ref{tab:table1}, we take three semantic primitives as an example. There is no relationship between the same primitives, so 'empty' is used to represent this relation. Relative to primitive 2, the relation between primitive 1 and primitive 2 is down. In the same way,  Relative to primitive 1, the relation between primitive 2 and primitive 1 is up. Similar relations exist among other primitives.

\subsection{Semantic relation template: the definition of semantic tree}
In the process of building semantic tree, the semantic relation template is an extremely essential part. It describes the relation distribution of the upper semantic with a specific class. We use a statistical method to build semantic relation template of the corresponding category. And the template is expressed by
\begin{equation}
\label{con:f2}
\begin{split}
\textbf{T}=(\sum_{k=1}^K r_1(k),\sum_{k=1}^K r_2(k), \cdots, \sum_{k=1}^Kr_\alpha(k))\\
\alpha = \alpha_{\text{int}}+\alpha_{\text{dis}}+\alpha_{\text{ang}}+\alpha_{\text{loc}}\\
\end{split}
\end{equation}
where $k$ represents the $k$-th training sample ($k=1, 2, \cdots, K$). $\alpha$ represents the  number of relations in the semantic relation table $\textbf R$. $r_i(k)$ represents the $i$-th relation value of $k$-th training sample.

As for classification task with $N$ classes, we denote $\textbf T_j$as the $j$-th semantic relation template which corresponds to the $j$-th class. For an input image, the semantic relation table $\textbf R$ can be calculated through Eq.~\ref{con:f1}. We adopt the product operation to compute the matching degree $d_j$ between the $\textbf R$ and $\textbf T_j$, which can be denoted by
\begin{equation}
\label{con:f3}
d_j = \frac1{{\|\textbf T_j\|}_1}(\textbf T_j\cdot \textbf{R})
\end{equation}

Then, we adopt the $softmax$ function to generate the probability $p_j$ of the corresponding category, which can be expressed as
\begin{equation}
\label{con:f4}
p_j = \frac{exp(d_j)}{\sum_{j=1}^Nexp(d_j)}
\end{equation}

\subsection{Loss function of the SCN}
Both semantic tree and the CapsNet modules have a prediction probability vector with $N$ dimensions, which can be represented by $({{p}_{1}},{{p}_{2}}...{{p}_{N}})$ and $({{q}_{1}},{{q}_{2}}...{{q}_{N}})$ respectively.
We design a linear function $f\left( \cdot  \right)$ to get the final output probability ${o}_{i}$:
\begin{equation}
\label{con:f5}
{{o}_{i}}=f({{p}_{j}},{{q}_{i}})={{\beta }_{1}}{{p}_{j}}+{{\beta }_{2}}{{q}_{j}}
\end{equation}
where ${{\beta }_{1}}, {{\beta }_{2}}$  represent the fusion coefficients of two modules respectively.

The $j$-th margin loss functions of the semantic tree, the CapsNet module and the SCN are expressed by
\begin{equation}
\begin{split}
\label{con:f6}
&L_{j}^{st}={{H}_{j}}\max {{(0,0.9-{{p}_{j}})}^{2}}+0.5\times (1-{{H}_{j}})\max {{(0,{{p}_{j}}-0.1)}^{2}}\\
&L_{j}^{caps}={{H}_{j}}\max {{(0,0.9-{{q}_{j}})}^{2}}+0.5\times (1-{{H}_{j}})\max {{(0,{{q}_{j}}-0.1)}^{2}}\\
&L_{j}^{scn}={{H}_{j}}\max {{(0,0.9-{{o}_{j}})}^{2}}+0.5\times (1-{{H}_{j}})\max {{(0,{{o}_{j}}-0.1)}^{2}}
\end{split}
\end{equation}

Their loss functions are represented respectively by
\begin{equation}
\begin{split}
\label{con:f7}
L^{st}=\frac{1}{M}\sum\limits_{m=1}^{M}{(\sum\limits_{j=1}^{N}{{L_{j}^{st}}})}\\
L^{caps}=\frac{1}{M}\sum\limits_{m=1}^{M}{(\sum\limits_{j=1}^{N}{{L_{j}^{caps}}})}\\
L^{scn}=\frac{1}{M}\sum\limits_{m=1}^{M}{(\sum\limits_{j=1}^{N}{{L_{j}^{scn}}})}
\end{split}
\end{equation}

In order to enable the semantic tree to guide the CapsNet to learn the representation of indescribable features, we design a total loss function $L$. That is
\begin{equation}
\begin{split}
&L=\frac{1}{1+{{e}^{100(0.5-\tau)}}}{{L}^{scn}},~~~
\tau=\frac{{{L}^{st}}}{{{L}^{st}}+{{L}^{caps}}}
\label{con:f8}
\end{split}
\end{equation}
where $H_j$ is the one-hot vector of the true label,  which belongs to $\{0,1\}$. $M$ is the batch size of training images. $N$ represents the number of categories. $\tau$ is the proportion of ${L}^{st}$ in the sum of ${L}^{st}$ and ${L}^{caps}$. Especially, Eq.~\ref{con:f8} has some special properties:
\begin{itemize}
\item[a)] It is a smooth loss function and can adjust the weights between the semantic tree and the CapsNet adaptively.
\item[b)] When ${L}^{st}$ is much larger than ${L}^{caps}$, $\tau$ is close to 1 and loss function ${L}$ is also close to ${L}^{scn}$. In the same way,  when ${L}^{st}$ is much smaller than ${L}^{caps}$, $\tau$ is close to zero and loss function ${L}$ is also close to zero.

\end{itemize}

\section{Experimental Evaluation}
We evaluate the performance of the SCN on MNIST, GTSRB datasets and corresponding small datasets through a series of experiments and compare our SCN with the original data-driven CapsNet both qualitatively and quantitatively. Furthermore we test the adversarial robustness of SCN on the adversarial test set produced by different adversarial attacks.

\subsection{Semantic tree of MNIST and GTSRB}
The MNIST consists of 28$\times$28 grayscale images of handwritten digits with 60000 training images and 10000 test images. We make some small training sets including 10000, 5000, 4000, 3000 and 2000 training images. The number of test set images keeps 10000. And the GTSRB dataset consists of real-world images and is much more complex, which contains 34799 training images and 12630 test images. In the same way, we also make some small training sets such as 0.9, 0.8, 0.7, 0.6, 0.5 times of the original training set.

For MNIST dataset, we conduct the semantic primitive detection on MNIST with the template matching method. Because all dataset are grayscale images, there is no need to take the color primitive into consideration. There are the following three steps during the semantic primitives detection for the MNIST.

\textbf{Image thinning:} Firstly, we perform an image thinning process in order to achieve the image skeletonization, which makes our semantic primitives features more prominent.

\textbf{Shape Detection:} In this step, we search for closed contours in the thinned image. The closed contours include circle or ellipse primitives. If there are closed contours in the MNIST image, such as number 0, 6, 8 and 9, we remove them from the thinned image. Otherwise, the thinned image remains unchanged. We also detect other semantic primitives in the output images of contour detecting. Here, we set another two kinds of semantic primitives: lines and arcs. So there are three types of semantic primitives for MNIST in total: closed contours (circle and ellipse), lines and arcs with different angles, size, length and so on. Eventually, for every image, we choose the two most typical primitives of each type.

Through the above three steps, we obtain all semantic primitives for MNIST. Each semantic primitive has a set of properties represented by an eight-dimensional vector. The properties of these semantic primitives are very important because they are used to make up the digit instance. For the closed contour, we use center point coordinates, length of the long and short axis and rotation angle of the center in horizontal direction as the properties. For the line, we use starting point and end point coordinates, length and angle with horizontal axis as the properties. For the arc, we use starting point, middle point and end point coordinates as the properties.

As for GTSRB dataset, We firstly extract color and shape primitive from the GTSRB dataset, which mainly contains triangle, circle, red, blue and so on. Then, we adopt the template matching method to detect the symbol primitives . There are the following several steps during the detection for the GTSRB.

\textbf{Image preprocessing:} Influenced by the change of illumination, the traffic sign images acquired in natural background have low brightness and contrast. In order to relieve these problems, histogram equalization method is used to improve the quality of original images. What is more, we obtain color semantic primitives through color segmentation.

\textbf{Shape detection:} After the previous operation, we take the template matching method to detect the shape primitives and symbol primitives of the dataset such as circle, triangle, bicycles, pedestrians and so on. Especially, adaptive two valued method is used to keep details in the image.

After detecting all semantic primitives for MNIST, we obtain $\alpha_1=8$ intersect features $\textbf R ^{\text{int}}$, $\alpha_2=95$ distance features $\textbf R ^{\text{dis}}$, $\alpha_3=13$ location relation features $\textbf R ^{\text{loc}}$, $\alpha_4=11$ included-angle relation features $\textbf R ^{\text{ang}}$. Thus, we can obtain corresponding semantic relation table $\textbf R$ with 127 dimensions.  For GTSRB dataset, we obtain 3 shape semantic primitives, 3 color semantic primitives and 43 internal primitives of GTSRB images. Furthermore, to acquire the semantic relation template $\textbf T_j$ of the $j$-th category, we perform a cumulative operation on semantic relation tables of training images. For each image in the test set, the probability value of every category images can be easily computed by Eq.\eqref{con:f3} and Eq.\eqref{con:f4}. The category with  max probability value is the semantic tree's classification result. Eventually, we obtain the $63\%$ accuracy on MNIST and GTSRB datasets. To integrate the semantic tree and the assisted CapsNet, we set the fusion coefficient of $\beta_1,\beta_2$  as 0.6 and 0.4 in Eq.\eqref{con:f5}.

\subsection{Training}
In the SCN,  the CapsNet module uses back propagation and dynamic routing algorithms to update network parameters and the parameters of the semantic tree module are predefined and remain unchanged. Furthermore, we use the Adam optimizer to minimize the loss function in Eq.\ref{con:f8} with gradually decline learning rate.

\subsection{Visualization and understanding of semantic computing network}
\begin{figure}[!ht]
\centering
\subfigure[256 feature maps in the original CapsNet]{
    \includegraphics[height=2.2in,width=2.2in]{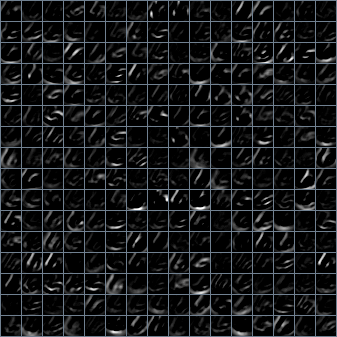}}
\subfigure[256 feature maps in SCN]{
    \includegraphics[height=2.2in,width=2.2in] {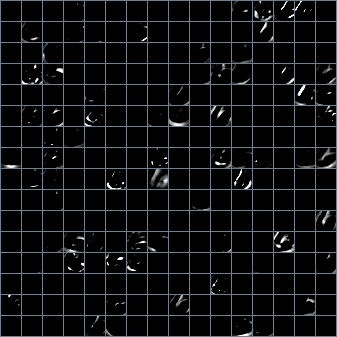}}
\subfigure[128 feature maps in SCN]{
    \includegraphics[height=1.5in,width=3in]{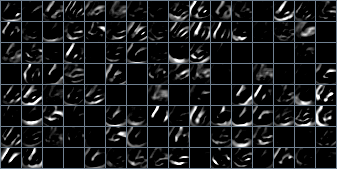}}
\subfigure[64 feature maps in SCN]{
    \includegraphics[height=1.5in,width=1.5in] {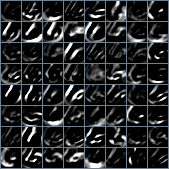}}
\caption{The difference of the feature maps between the original CapsNet and the proposed SCN. And the black pixels of feature maps represent that their values are negative or zero.}
\label{fig4}
\end{figure}

In this section, we perform some visualization experiments between the SCN and the original CapsNet. The difference between their feature maps in the first layer is shown in Fig. \ref{fig4}. Here, we take the digit  '6' of the MNIST dataset as an example.

As shown in Fig. \ref{fig4}(b), it is obvious that for the SCN there are many invalid feature maps because their pixels are all zero, while for the CapsNet all feature maps include varied edges, texture information (Fig. \ref{fig4}(a)). In our experiments, it is interesting that the number of black invalid feature maps becomes smaller after reducing the channels number of CapsNet module in the SCN to 128 and 64 respectively (Fig. \ref{fig4}(c, d)).  In particular, the 64 feature maps in first layer of the CapsNet module in the SCN have no black ones. Most important of all, we find the feature maps in the SCN (Fig. \ref{fig4}(b)) are the remaining parts after the original feature maps(Fig. \ref{fig4}(a)) remove the semantic primitives. These remaining parts are learned through the CapsNet. The same phenomena appear in the PrimaryCaps layer.  In other words, our SCN does not need too many filters. We consider simplifying our SCN according to the visualization results.

\subsection{ Experiments on simplified SCN}
\begin{figure}[H]
  \centering
  \includegraphics [height=5cm,width=8cm] {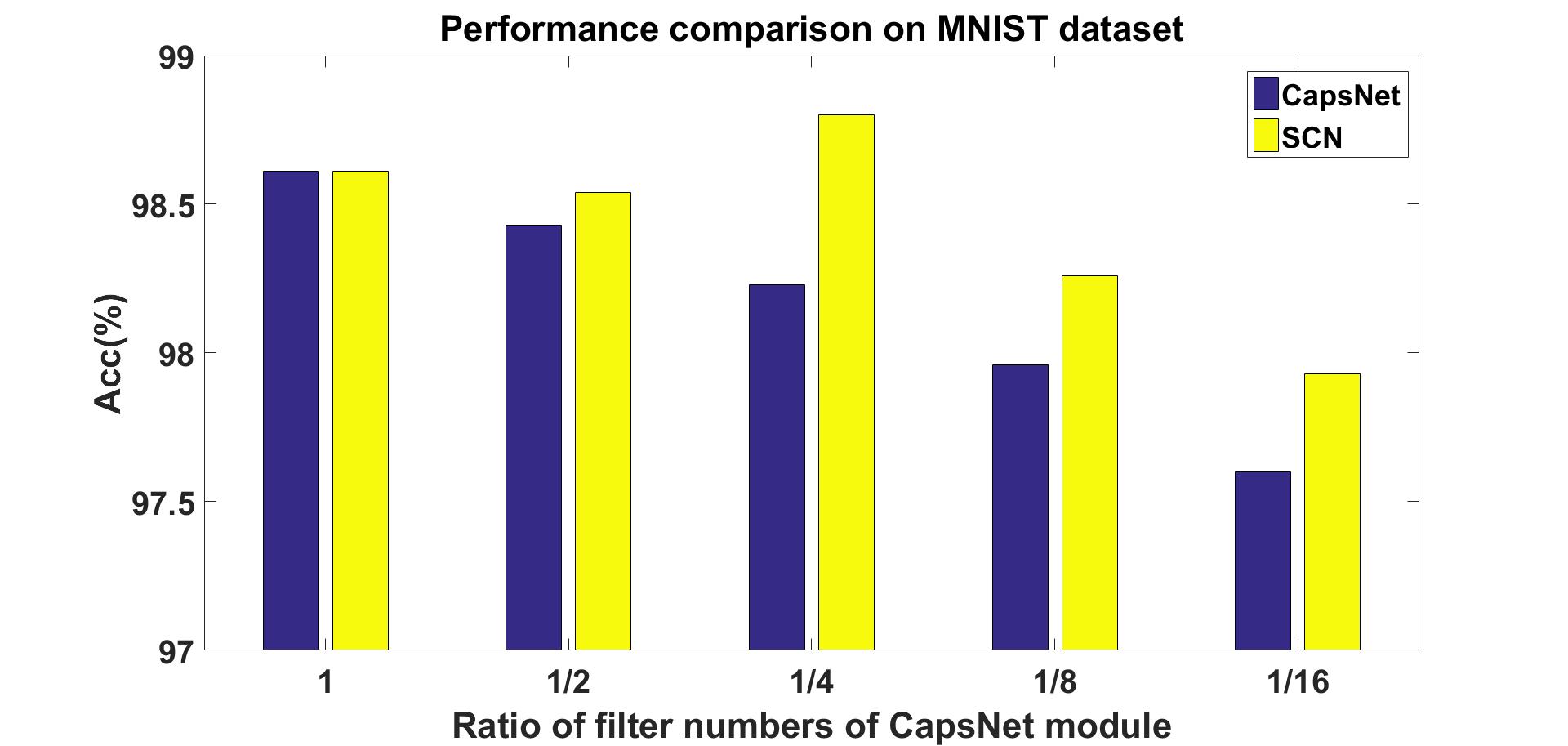}\\
  \caption{ Test accuracy of different networks.}
  \label{fig5}
\end{figure}

\begin{figure}[H]
  \centering
  \includegraphics [height=5cm,width=8cm] {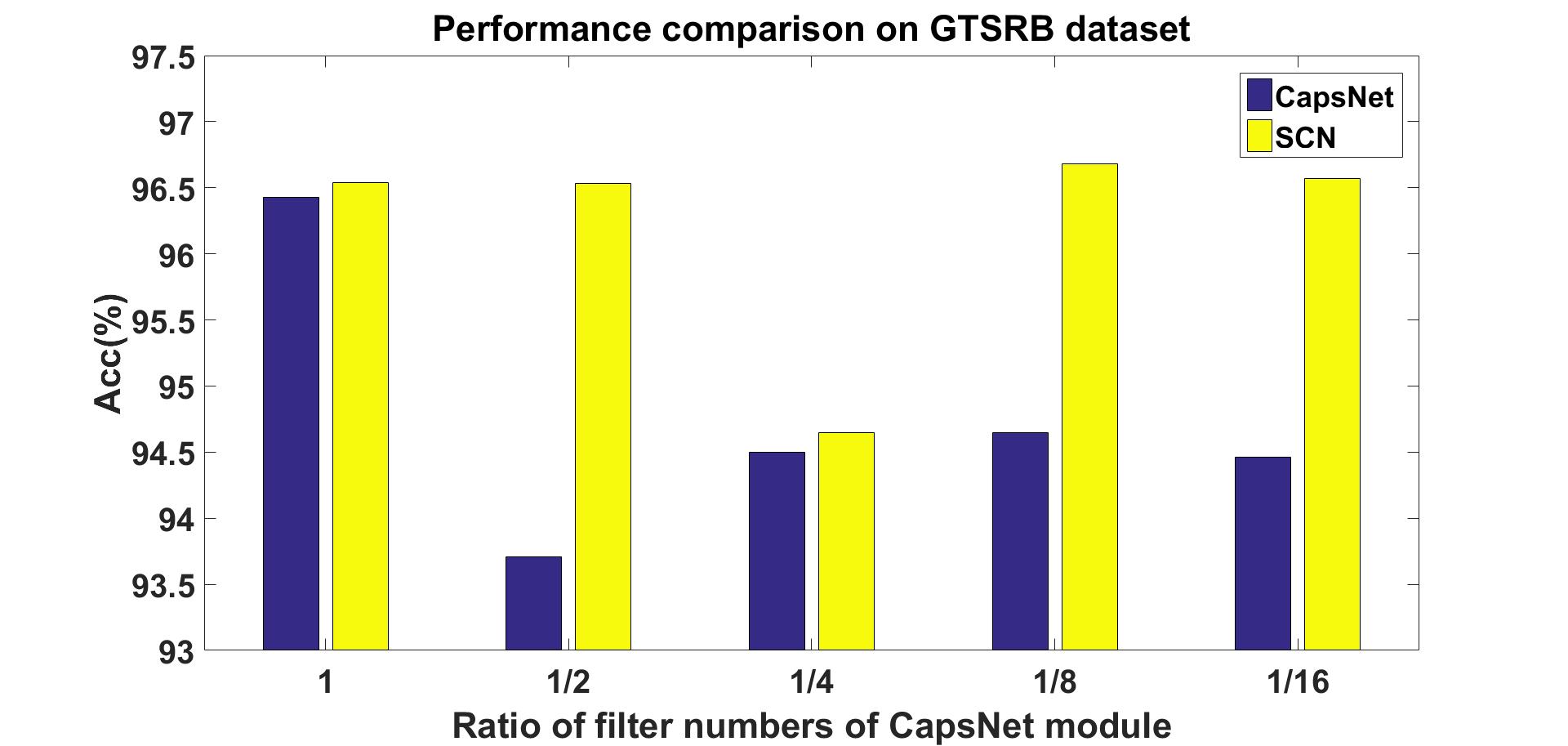}\\
  \caption{ Test accuracy of different networks.}
  \label{fig6}
\end{figure}

According to section 5.3, some experiments are designed to evaluate the performance of the simplified SCN. When simplifying the SCN, we reduce the filter numbers of both first layer and PrimaryCaps layer in the CapsNet module to its 1/2, 1/4, 1/8, 1/16, noted by SCN(1/2), SCN(1/4), SCN(1/8), SCN(1/16). The accuracies of these five pairs of SCNs and CapsNet networks on MNIST and GTSRB datasets are shown in Fig. \ref{fig5}. \ref{fig6}.

As shown in Fig. \ref{fig5} .\ref{fig6}, it is obvious that the SCN outperforms the CapsNet in all comparison experiments. Especially, our network obtains the best result at SCN(1/4) and SCN(1/8) on the MNIST and GTSRB datasets. On the MNIST dataset, when we inspect its feature maps on the digit'6' (Fig. \ref{fig4}(d)), SCN(1/4) has almost no invalid feature maps in the first layer, which means that SCN(1/4) fully utilizes all convolutional filters. The same phenomenon appears on the GTSRB dataset. SCN(1/8) fully utilizes all convolutional filters. Through simplifying, the SCN also reduces time and space complexity. The related experiment results are shown in Table. \ref{tab:table2}.

\begin{table*}[ht]
\centering
\caption{Time and space complexity comparison}
\label{tab:table2}
  \begin{tabular}{cccccccc}
    \toprule
    Datasets & Complexities & CapsNet & SCN & SCN(1/2) & SCN(1/4) & SCN(1/8) & SCN(1/16) \\
    \midrule
    \multirow{2}*{GTSRB} & Time & $2\times{{10}^{8}}$ & $2\times{{10}^{8}}$ & $6.4\times{{10}^{7}}$ & $\textbf{$2.2\times{{10}^{7}}$}$ & $8.5\times{{10}^{6}}$ & $3.6\times{{10}^{6}}$ \\
      {}  & Space & $3.7\times{{10}^{7}}$ & $3.7\times{{10}^{7}}$ & $1.8\times{{10}^{7}}$ & $\textbf{$8.9\times{{10}^{6}}$}$ & $4.4\times{{10}^{6}}$ & $2.1\times{{10}^{6}}$ \\
    \hline
    \multirow{2}*{MNIST} & Time & $2\times{{10}^{8}}$ & $2\times{{10}^{8}}$ & $5.2\times{{10}^{7}}$ & $\textbf{$1.4\times{{10}^{7}}$}$ & $4.2\times{{10}^{6}}$ & $1.4\times{{10}^{6}}$ \\
        {} & Space & $6.8\times{{10}^{6}}$ & $6.8\times{{10}^{6}}$ & $2.1\times{{10}^{6}}$ & $\textbf{$7.1\times{{10}^{5}}$}$ & $2.7\times{{10}^{5}}$ & $1.1\times{{10}^{5}}$ \\
    \bottomrule
  \end{tabular}
\end{table*}

As shown in  Table \ref{tab:table2}, along with the proposed SCN simplifies, its time and space complexities are also reduced. Especially, on the GTSRB dataset, the time complexity of SCN(1/8) is only 0.04 times that of the CapsNet. The same on the MNIST dataset, the time complexity of SCN(1/4) is only 0.07 times that of the CapsNet.

\begin{figure}[ht]
  \centering
  \includegraphics [height=5cm,width=8cm] {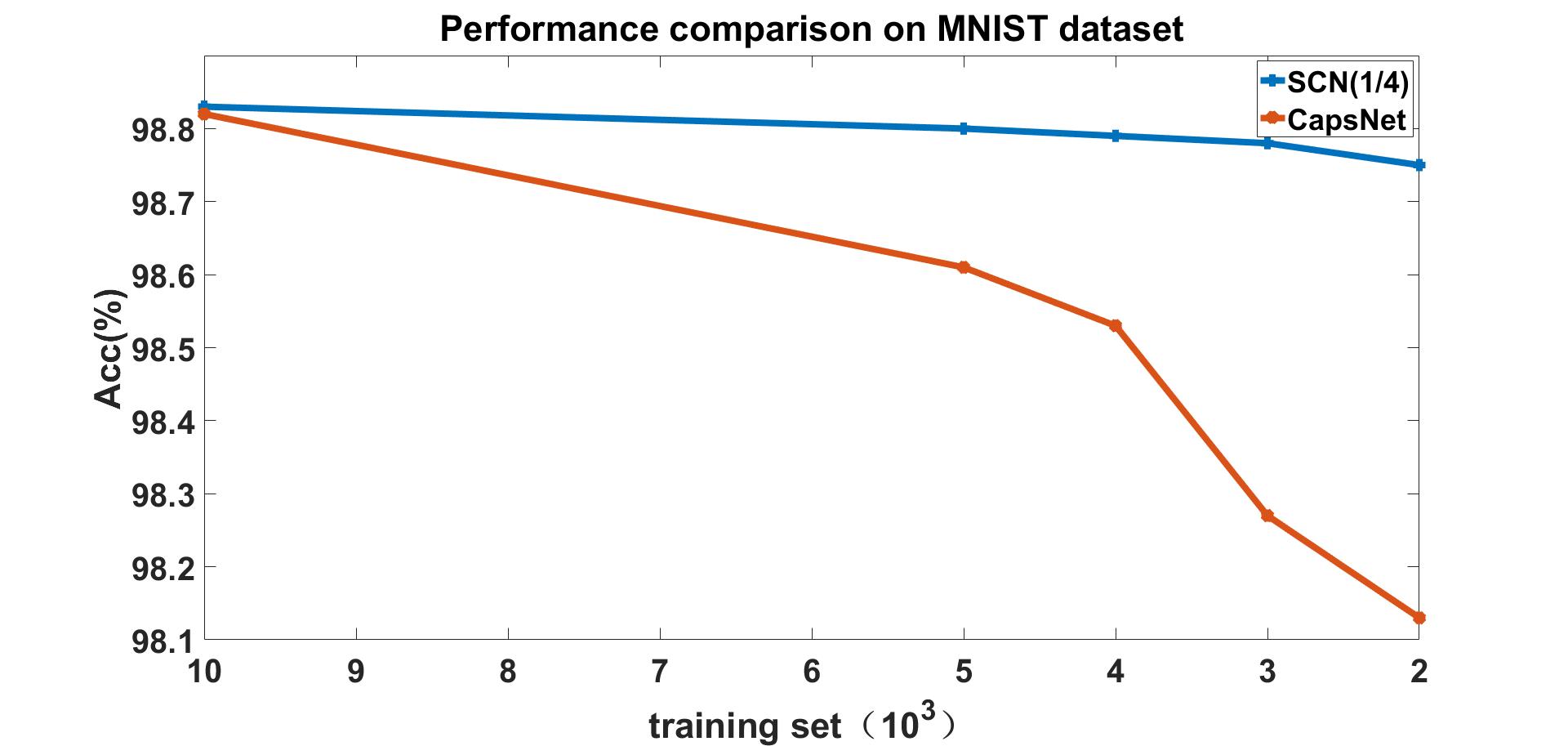}\\
  \caption{  Performance comparison among different size training sets.}
  \label{fig7}
\end{figure}

\begin{figure}[ht]
  \centering
  \includegraphics [height=5cm,width=8cm] {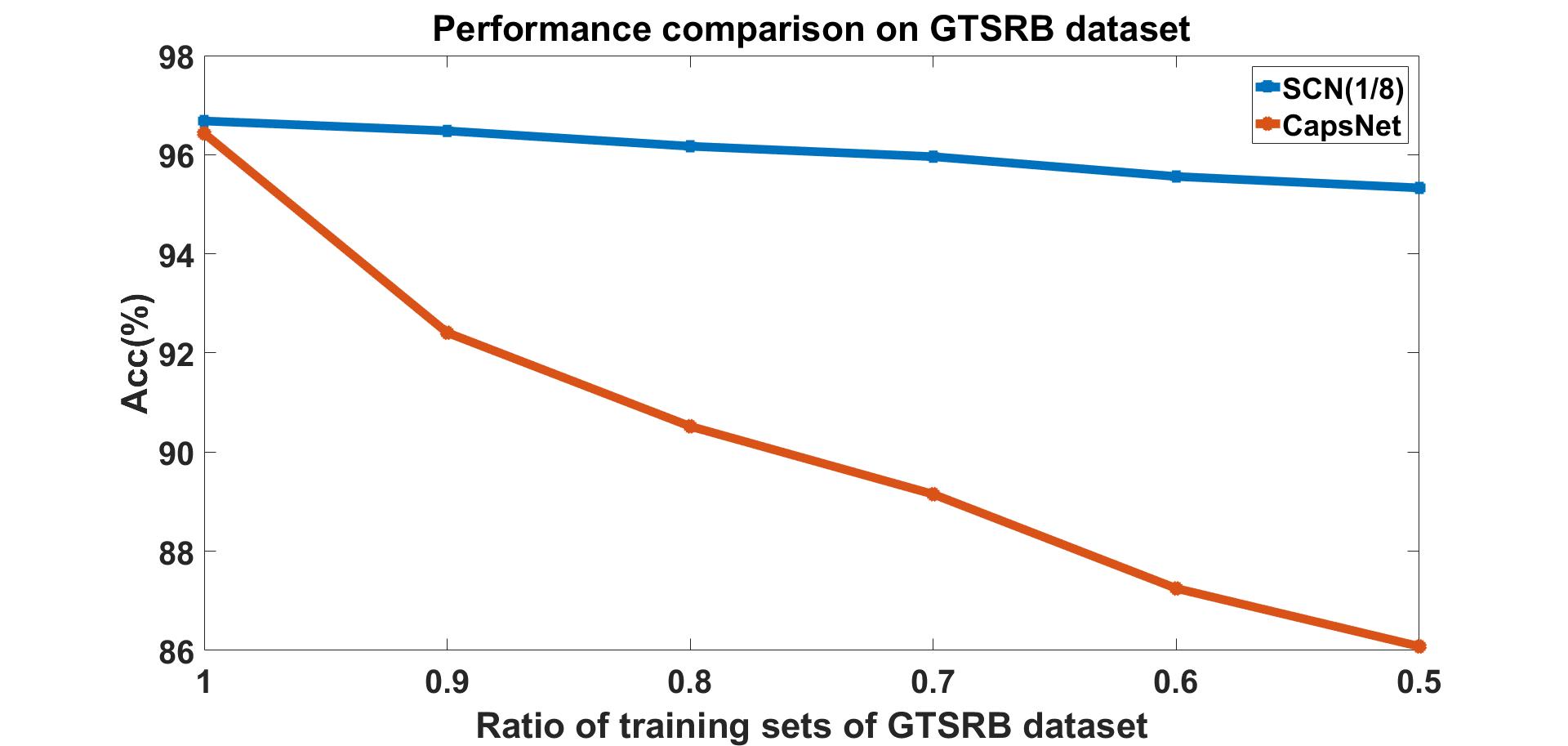}\\
  \caption{  Performance comparison among different size training sets.}
  \label{fig8}
\end{figure}

To further confirm that the SCN also has excellent performance on smaller datasets, we make different experiments on 10000, 5000, 4000, 3000 and 2000 training images on the MNIST dataset respectively. The number of test images always keeps 10000. We compare the original CapsNet and SCN(1/4) on these smaller training set. For GTSRB dataset, we use 1 times, 0.9 times, 0.8 times, 0.7 times, 0.6 times and 0.5 times training set to train the model. In particular, we compare the original CapsNet with SCN(1/8) on these smaller training set. And the accuracy curves on different training sets are shown in Fig. \ref{fig7}, \ref{fig8}.

As shown in Fig. \ref{fig7}, compared with the CapsNet, our SCN(1/4) obtains high and stable accuracy among different size training sets. When we reduce the training set from 10000 to 2000, the accuracy of the original CapsNet decreases sharply while the accuracy of SCN(1/4) almost remains unchanged. In the same way, Fig. \ref{fig8} shows our SCN(1/8) achieves a better performance than the original CapsNet at all different size training sets by large margin. Especially, SCN(1/8) obtains almost 10\% improvement than CapsNet at half of the training set. In other words, the proposed SCN can achieve better performance on fewer training samples with the help of the prior knowledge included in the semantic tree.

\subsection{ Adversarial Robustness Experiments}

Recently, there is a trend to study the robustness of traditional neural network to adversarial attacks. The adversarial examples can be recognized by humans, but they can fool the neural networks to make wrong classification results. Up to now, there are a variety of methods to create the adversarial examples such as FGSM ~\cite{Goodfellow2014Explaining} and BIM ~\cite{DBLP:journals/corr/KurakinGB16}. These ways have shown a great obstacle to convolutional neural networks on the image classification tasks.

\begin{figure}[ht]
  \centering
  \includegraphics [height=5cm,width=8cm] {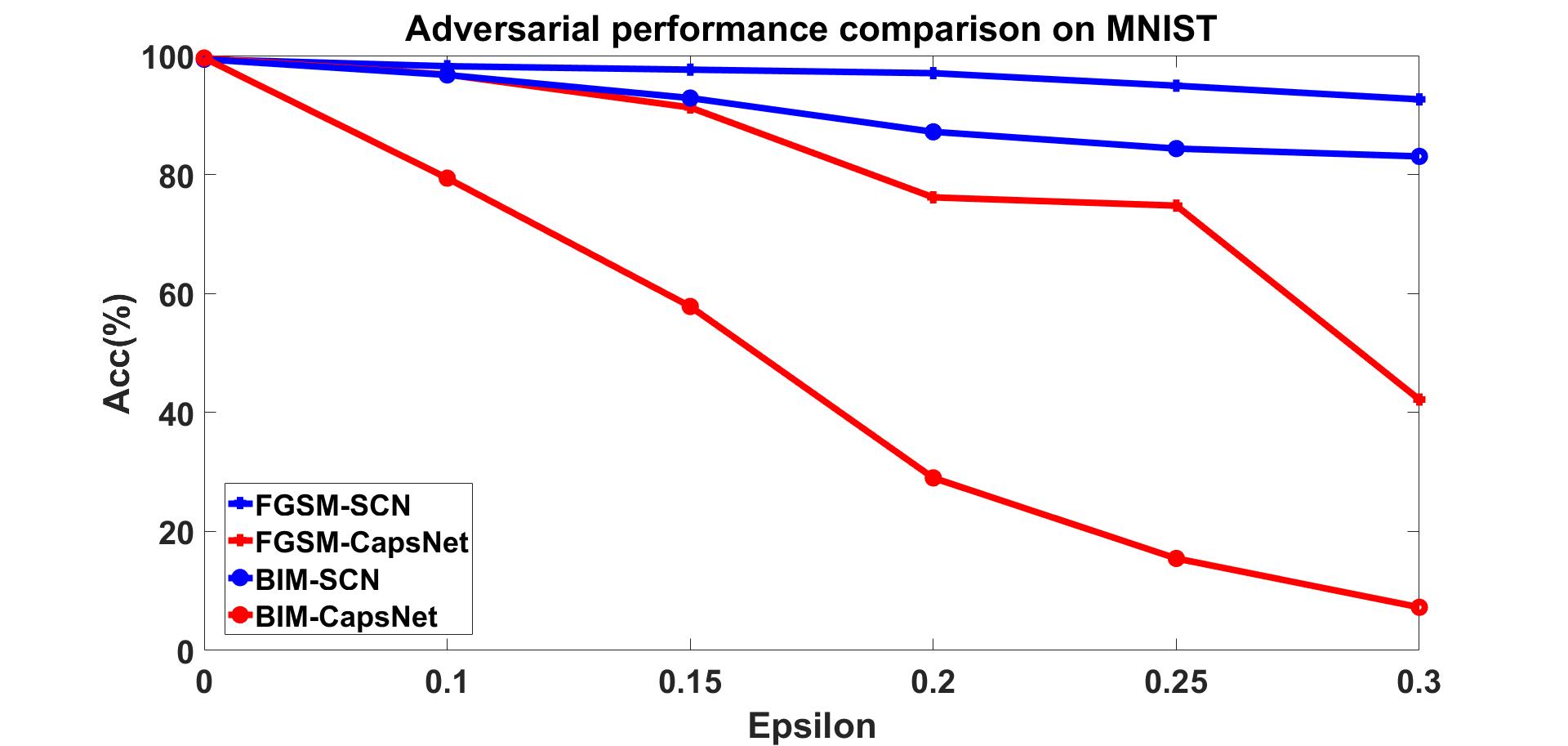}\\
  \caption{  Performance comparison among different adversarial samples of MNIST.}
  \label{fig9}
\end{figure}

\begin{figure}[ht]
  \centering
  \includegraphics [height=5cm,width=8cm] {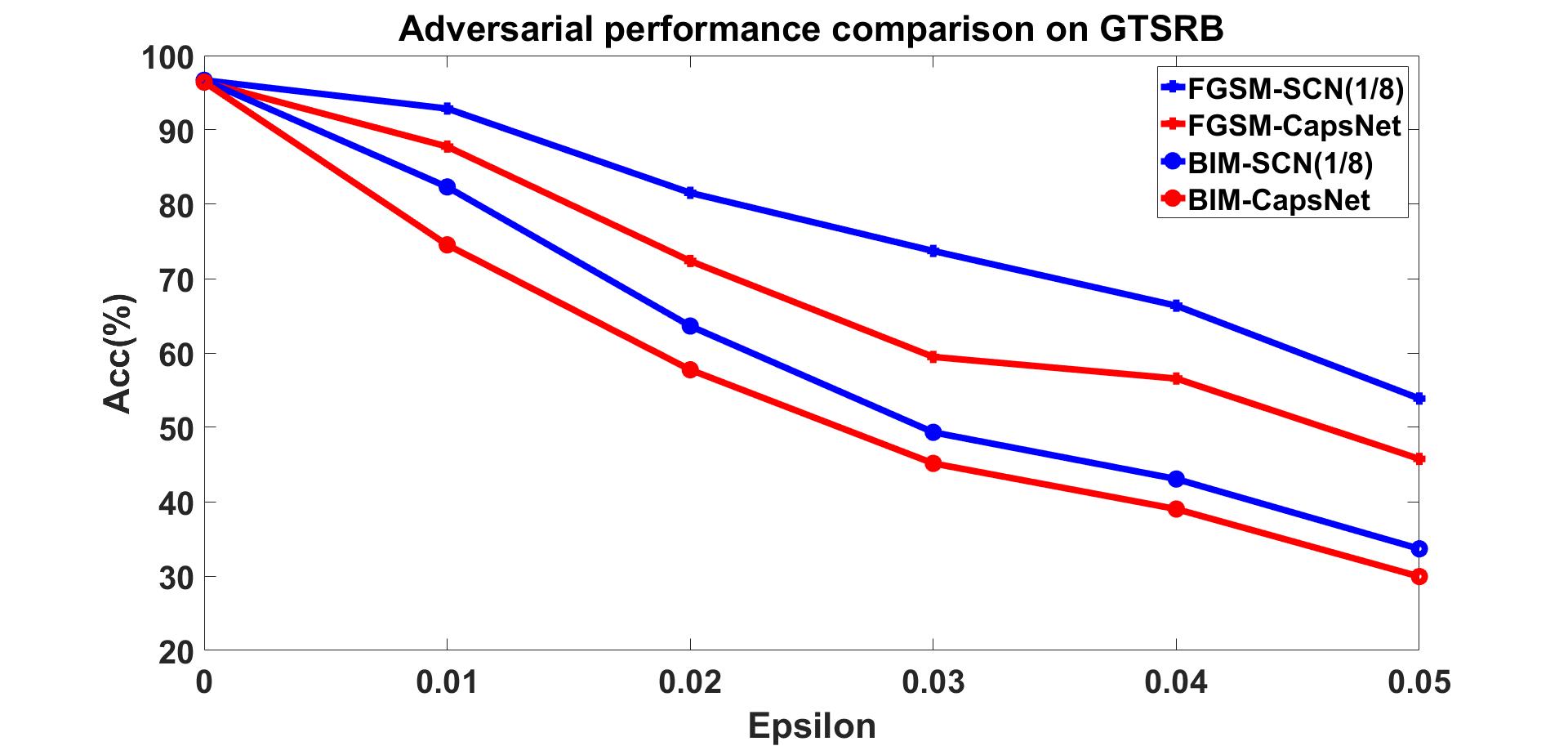}\\
  \caption{  Performance comparison among different adversarial samples of GTSRB.}
  \label{fig10}
\end{figure}

In order to confirm the robustness of the proposed SCN. we generate adversarial samples from the test set (MNIST and GTSRB datasets) using FGSM method and more complex adversarial attacks of the Basic Iterative Method. These adversarial samples can vary according to different perturbations (epsilon) and number of iterations. In this paper, For MNIST dataset, we add five levels of epsilon corresponding to eps=0.1, 0.15, 0.2, 0.25 and 0.3. For GTSRB dataset, we add five levels of epsilon corresponding to eps=0.01, 0.02, 0.03, 0.04 and 0.05. Especially, the number of iterations on the MNIST and GTSRB BIM adversarial samples are 2. The related experiment results are in Fig. \ref{fig9}. \ref{fig10}.

As shown in Fig. \ref{fig9}, it is obvious that our SCN is significantly more robust to adversarial attacks than the CapsNet at all adversarial samples. Our SCN achieves high and stable accuracy with the increase of eps on the FGSM attack. Especially at BIM eps=0.3, our model obtains over 75\% improvement. Most important of all, the CapsNet is trained on the 60000 training images, while SCN is only trained on the 5000 training images. The same phenomenon appears on GTSRB dataset (Fig. \ref{fig10}). Compared with the CapsNet, our SCN(1/8) achieves obvious excellent accuracy. Especially on the FGSM eps=0.03, our SCN(1/8) obtains over 14\% improvement. Most important of all, SCN(1/8) is a greatly simplified semantic compute network, whose time complexity is only 0.04 times that of the SCN.

\begin{table*}[!ht]
\centering
\caption{Adversarial experiments comparison on MNIST dataset}
\label{tab:table3}
  \begin{tabular}{cccc}
    \toprule
    Adversarial methods & CNN & CapsNet & SCN(training set = 5000) \\
    \midrule
     FGSM eps=0.1 & $89.13\%$ & $96.80\%$ & $\textbf{98.21}\%$ \\
    \hline
    FGSM eps=0.15 & $71.93\%$ & $91.27\%$ & $\textbf{97.63}\%$ \\
    \hline
    FGSM eps=0.2 & $41.12\%$ & $76.16\%$ & $\textbf{97.04}\%$ \\
    \hline
    FGSM eps=0.25 & $25.6\%$ & $74.77\%$ & $\textbf{94.93}\%$ \\
    \hline
    FGSM eps=0.3 & $14.62\%$ & $42.17\%$ & $\textbf{92.64}\%$ \\
    \hline
    BIM eps=0.1 & $21.81\%$ & $79.39\%$ & $\textbf{96.74}\%$ \\
    \hline
    BIM eps=0.15 & $1.98\%$ & $57.82\%$ & $\textbf{92.86}\%$ \\
    \hline
    BIM eps=0.2 & $0.82\%$ & $28.00\%$ & $\textbf{87.18}\%$ \\
    \hline
    BIM eps=0.25 & $0.66\%$ & $15.46\%$ & $\textbf{84.36}\%$ \\
    \hline
    BIM eps=0.3 & $0.61\%$ & $7.25\%$ & $\textbf{83.05}\%$ \\
    \bottomrule
  \end{tabular}
\end{table*}

\begin{table*}[!ht]
\centering
\caption{Adversarial experiments comparison on GTSRB dataset}
\label{tab:table4}
  \begin{tabular}{cccc}
    \toprule
    Adversarial methods & CNN & CapsNet & SCN(1/8) \\
    \midrule
     FGSM eps=0.01 & $62.67\%$ & $87.76\%$ & $\textbf{92.87}\%$ \\
    \hline
    FGSM eps=0.02 & $40.92\%$ & $72.37\%$ & $\textbf{81.54}\%$ \\
    \hline
    FGSM eps=0.03 & $29.26\%$ & $59.47\%$ & $\textbf{73.72}\%$ \\
    \hline
    FGSM eps=0.04 & $25.21\%$ & $56.56\%$ & $\textbf{66.36}\%$ \\
    \hline
    FGSM eps=0.05 & $19.13\%$ & $45.75\%$ & $\textbf{53.87}\%$ \\
    \hline
    BIM eps=0.01 & $38.13\%$ & $74.54\%$ & $\textbf{82.33}\%$ \\
    \hline
    BIM eps=0.02 & $21.62\%$ & $57.74\%$ & $\textbf{63.62}\%$ \\
    \hline
    BIM eps=0.03 & $17.56\%$ & $45.15\%$ & $\textbf{49.33}\%$ \\
    \hline
    BIM eps=0.04 & $14.32\%$ & $39.01\%$ & $\textbf{43.05}\%$ \\
    \hline
    BIM eps=0.05 & $10.65\%$ & $29.93\%$ & $\textbf{33.66}\%$ \\
    \bottomrule
  \end{tabular}
\end{table*}

As shown in Table \ref{tab:table3}, on MNIST dataset, we find that our model is less vulnerable to both FGSM and BIM adversarial attacks. Especially on the BIM adversarial samples with eps=0.2, 0.25, 0.3. the traditional convolutional neural networks basically do not work at all. While our model still keeps a high accuracy rate. For the GTSRB dataset (Table \ref{tab:table4}), our SCN(1/8) has more robustness than traditional convolutional neural networks at all adversarial samples. Especially on the BIM adversarial samples with eps=0.01, our SCN(1/8) obtains over 44\% improvement than traditional neural network.

\begin{figure*}[!ht]
\centering
\subfigure[Adversarial samples of different methods and perturbations on GTSRB]{
    \includegraphics[height=1.4in,width=6.6in]{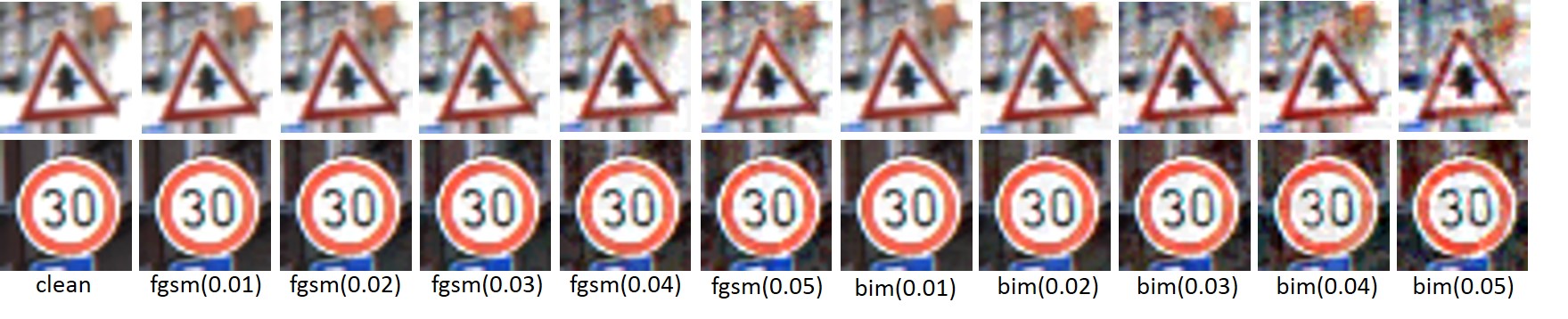}}
\subfigure[Adversarial samples of different methods and perturbations on MNIST]{
    \includegraphics[height=1.4in,width=6.6in] {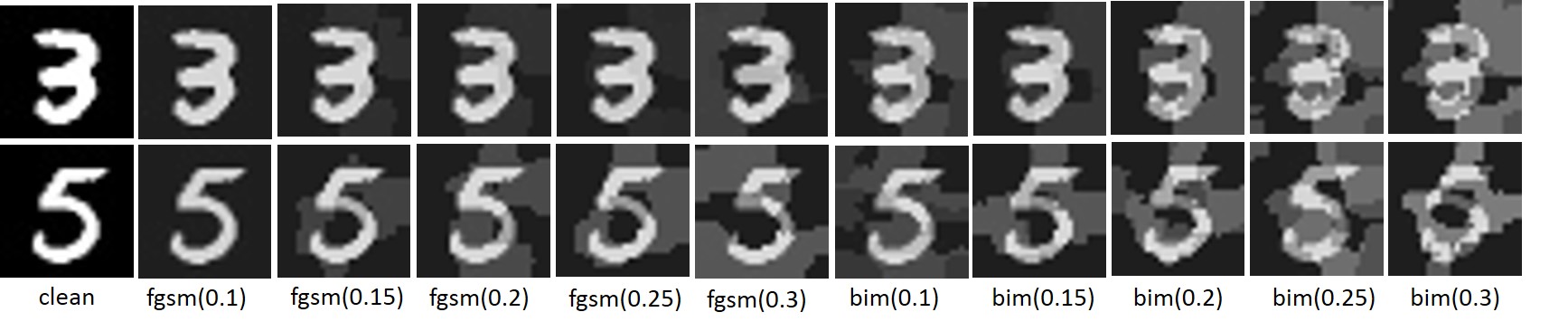}}
\caption{Comparison of adversarial samples resulting from different methods and perturbations. In both datasets clean images are classified correctly and all adversarial images are misclassified}
\label{fig11}
\end{figure*}

As shown in Fig. \ref{fig11}, it is obvious that human can easily identify the categories of all adversarial images. Therefore, our semantic tree is not directly affected by the adversarial attacks. While traditional neural networks become very vulnerable to these adversarial attacks. Furthermore, experimental results(Table \ref{tab:table3}, \ref{tab:table4}) explain that our SCN is much more robust to the adversarial samples than the traditional neural networks.

\begin{figure}[!ht]
  \centering
  \includegraphics [height=3in,width=3.6in] {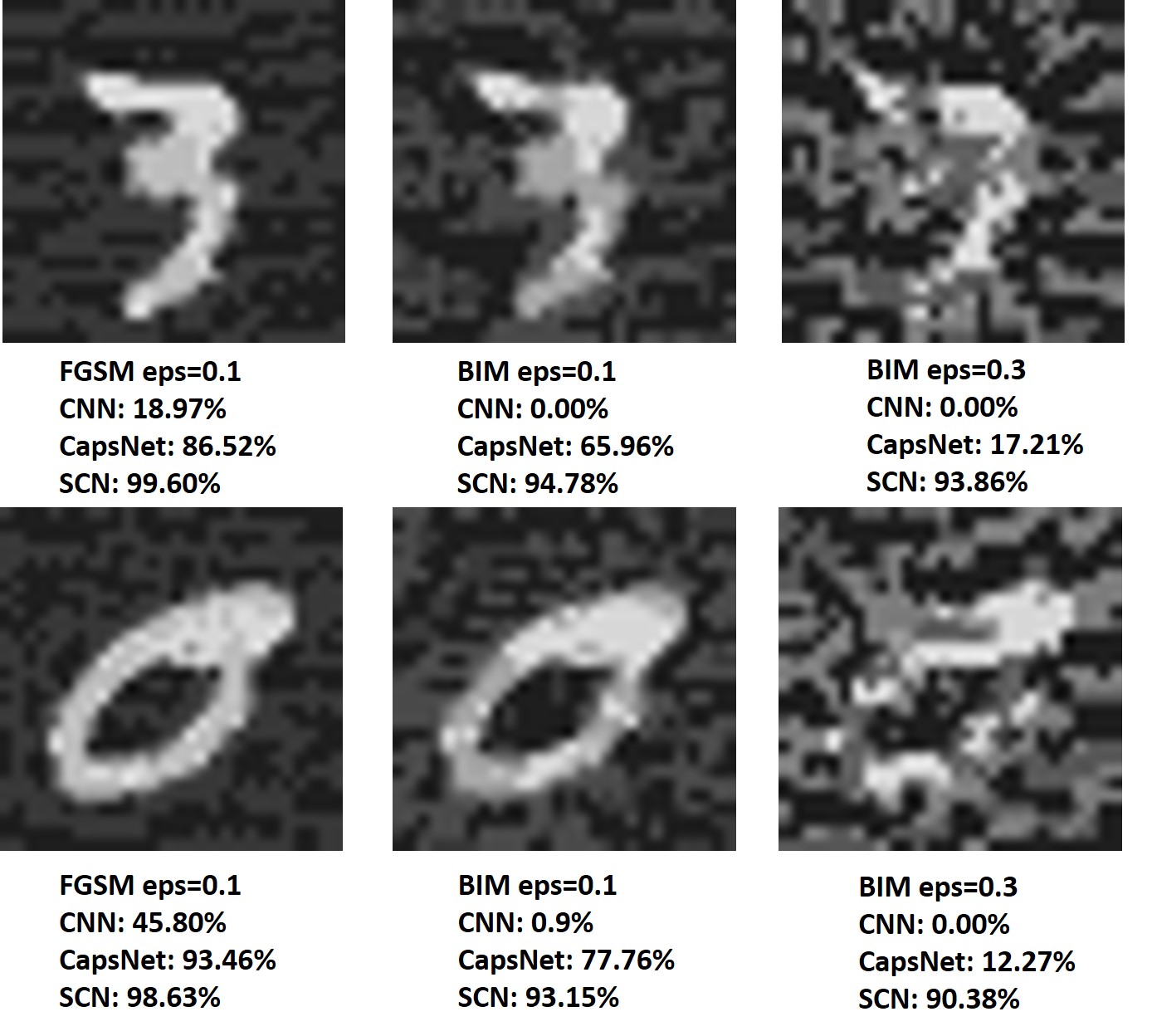}\\
  \caption{  Performance comparison of different networks on MNIST adversarial samples.}
  \label{fig12}
\end{figure}

\begin{figure}[!ht]
  \centering
  \includegraphics [height=3in,width=3.6in] {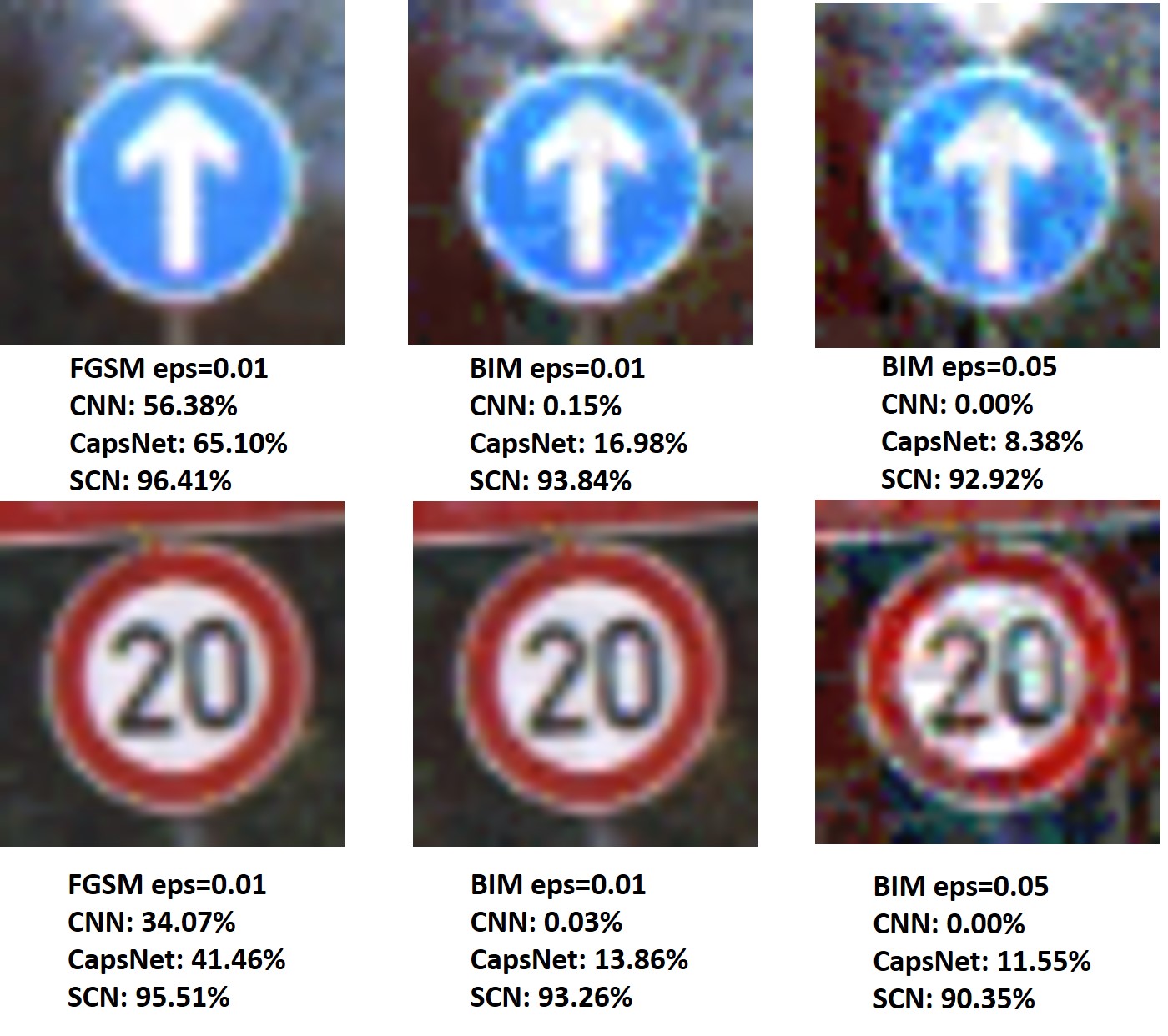}\\
  \caption{  Performance comparison of different networks on GTSRB adversarial samples.}
  \label{fig13}
\end{figure}

As shown in Fig. \ref{fig12}, we study the performance of different networks on the different adversarial samples. We select corresponding adversarial images on the FGSM eps=0.1, BIM eps=0.1 and BIM eps=0.3. With the deterioration of images quality, the accuracies of CNN and CapsNet decreased significantly. Due to the integrity of semantic information in adversarial images, our SCN can still maintain high confidence level. The similar phenomenon appears on the GTSRB dataset (Fig. \ref{fig13}), we select the images on the FGSM eps=0.01, BIM eps=0.01 and BIM eps=0.05.

\section{Conclusion and Future Work}
In this paper, we propose a novel knowledge-guided semantic computing network for different classification tasks, which includes two modules: a knowledge-guided semantic tree and a data-driven CapsNet. The semantic tree is pre-defined to describe the spatial structural relations of different semantics, we use the CapsNet to aid the semantic tree to learn indescribable features. The experimental results on MNIST and GTSRB datasets show that compared with the traditional data-driven network, our proposed semantic computing network can achieve better performance with fewer training samples and lower computational complexity. Especially, our SCN is much more robust to adversarial samples than traditional neural networks. Most of all, we propose a novel method that integrating prior knowledge into traditional neural networks for faster learning speed, fewer training samples and better interpretability.

We plan to evaluate our semantic computing network on more complex datasets, such as the ImageNet dataset. We also need to further optimize network architecture for better performance. The results presented in this paper motivate some researches about integration of human knowledge and traditional neural networks. We consider this as an important research direction for future work.
\ifCLASSOPTIONcaptionsoff
  \newpage
\fi

\bibliographystyle{unsrt}
\bibliography{Citation}

\end{document}